\documentclass[letterpaper, 10 pt, conference]{ieeeconf}  

\IEEEoverridecommandlockouts                              

\title{\LARGE \bf Soft Actor-Critic-based Control Barrier Adaptation for Robust Autonomous Navigation in Unknown Environments}

\author{Nicholas Mohammad, Nicola Bezzo
\thanks{Nicholas Mohammad and Nicola Bezzo are with the Autonomous Mobile Robots Lab (AMR Lab) and the Department of Electrical and Computer Engineering, University of Virginia, Charlottesville, VA 22903, USA 
        {\tt\small \{nm9ur, nbezzo\}@virginia.edu}}%
}%

\newcommand{\subparagraph}{}

\newcommand{\editcolor}{\color{black}}

\usepackage{graphicx}
\usepackage{epstopdf}
\usepackage{amsmath}
\usepackage{amssymb}
\usepackage{subfigure}
\usepackage{multirow}
\usepackage{pbox}
\usepackage{algorithm}
\usepackage{algpseudocode}
\usepackage{titlesec}
\usepackage{bm}
\usepackage{url}
\usepackage{dblfloatfix} 
\usepackage{booktabs} 
\usepackage{siunitx}  
\usepackage{float}
\usepackage[colorinlistoftodos]{todonotes}

\usepackage{algpseudocode,algorithm,algorithmicx}  
\algrenewcommand\algorithmicrequire{\textbf{Precondition:}}  
\algrenewcommand\algorithmicensure{\textbf{Postcondition:}}

\titleformat{\subsubsection}[runin] 
  {\itshape\normalsize}          
  {\arabic{subsubsection})}         
  {.5em}                             
  {}                                

\titlespacing*{\subsubsection}{1em}{0.5em}{.5em}

\newcommand{\norm}[1]{\left\lVert#1\right\rVert}

\renewcommand{\baselinestretch}{.94}

\usepackage{xcolor}

\DeclareMathOperator{\atantwo}{atan2}

\begin{document}

\graphicspath{ {./figs2/} }
\maketitle

\begin{abstract}
Motion planning failures during autonomous navigation often occur when safety constraints are either too conservative, leading to deadlocks, or too liberal, resulting in collisions. To improve robustness, a robot must dynamically adapt its safety constraints to ensure it reaches its goal while balancing safety and performance measures. To this end, we propose a Soft Actor-Critic (SAC)-based policy for adapting Control Barrier Function (CBF) constraint parameters at runtime, ensuring safe yet non-conservative motion. The proposed approach is designed for a general high-level motion planner, low-level controller, and target system model, and is trained in simulation only. Through extensive simulations and physical experiments, we demonstrate that our framework effectively adapts CBF constraints, enabling the robot to reach its final goal without compromising safety.


\end{abstract}

\section{Introduction} \label{sec:intro}

Autonomous mobile robots (AMRs) are becoming increasingly prevalent in industries such as inspections, search and rescue, and transportation. Despite their potential, safety assurance remains a major hurdle to widespread adoption. This challenge is highlighted by recent high-profile incidents, including Waymo's recall of self-driving vehicles due to collisions caused by navigation failures \cite{nhtsa2024waymo}. This safety problem was also evidenced at the 2024 ICRA BARN challenge \cite{xiao2024barn}, in which no team succeeded in navigating their robot through a series of complex, cluttered environments without collisions. These collisions are often attributed to failures in both high-level path planning and low-level tracking control, the former of which we investigated in our prior work \cite{mohammad2024planner}.

Addressing this challenge at its root requires the low-level controller to reason about surrounding obstacles and generate safety-minded control inputs, regardless of the high-level planner output. A promising approach for designing such a safety-critical controller is Control Barrier Functions (CBFs) \cite{ames2019cbf}, which act as a filter for the low-level controller to ensure system safety. While there are existing applications of CBFs to obstacle avoidance, \cite{koushil2022polycbf,li2023cbf}, the results are typically limited to simulations only and rely on a fixed parameter $\alpha$ which controls the desired level of conservatism for the safety filter.
However, the required conservatism typically varies across an environment, necessitating runtime adaptation of $\alpha$. 

\begin{figure}[t!]
    \centering
    \includegraphics[width=0.45\textwidth]{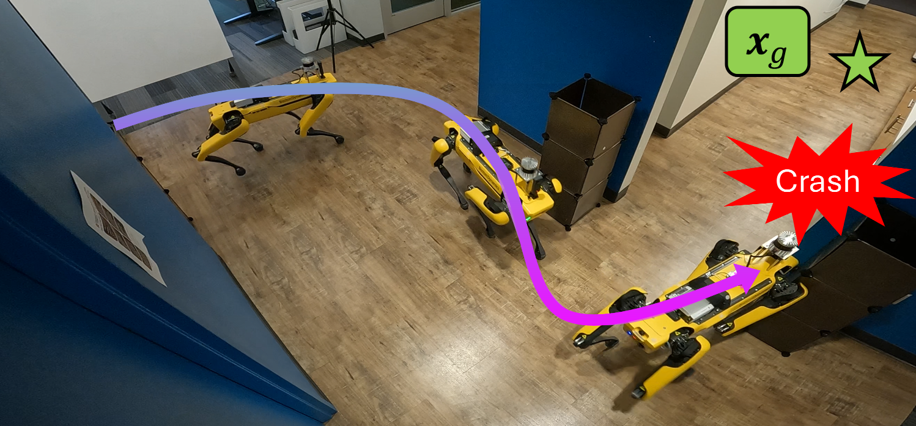} \\
    \vspace{+2pt}
    \includegraphics[width=0.45\textwidth]{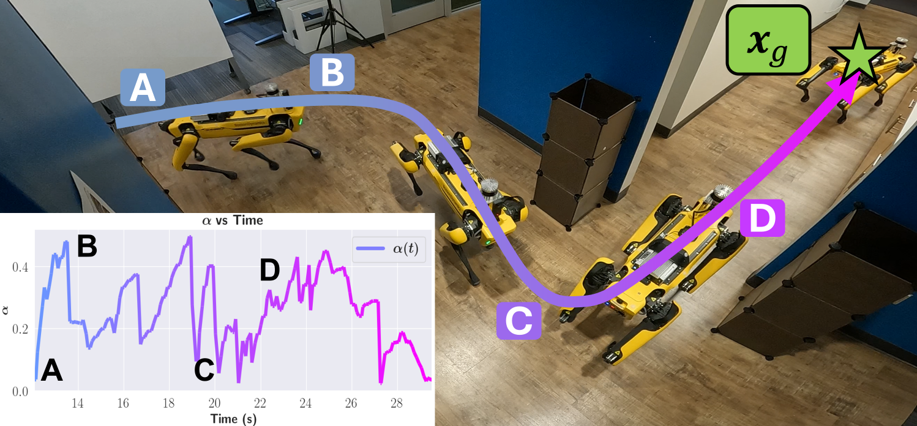}\label{fig:spot_with_cbf}
    \caption{(Top) Traditional motion planning framework fails to avoid obstacles when navigating a narrow corridor. (Bottom) Proposed framework adapts CBF constraints in order to reach the final goal while ensuring safety.}
    \vspace{-20pt}
    \label{fig:intro}
\end{figure}

For instance, consider the case in Fig.~\ref{fig:intro}, where the robot must navigate through a narrow, unknown environment. In the top image, without any safety scheme, the robot crashes because the low level controller fails to accurately track the high level reference. If a fixed, overly conservative CBF constraint is applied, the robot might become deadlocked at the entrance to the corridor (point C in Fig.~\ref{fig:intro}). However, by dynamically increasing the level of conservatism (lower $\alpha$) as it approaches the narrow passage, then relaxing the constraint (higher $\alpha$) once aligned (point D), the robot can safely negotiate the opening and reach its goal $\bm{x}_g$. This scenario is discussed further in Sec.~\ref{sec:experiment}. 

To enhance the robustness of the motion planning paradigm, we propose a data-driven framework that dynamically adjusts $\alpha$ at runtime based on the robot's state, desired input, and sensed obstacle configuration. For this, we leverage the Soft Actor-Critic (SAC) algorithm \cite{haarnoja2019sac}, which uses an entropy-based reward structure to learn a stable, stochastic policy $\pi_\theta$. While real-world applications of Reinforcement Learning (RL) are often restricted to simplified environments like OpenAI Gym \cite{brockman2016openaigym} for training, we develop a custom pipeline which trains the SAC policy alongside the low-level controller in a high-fidelity simulator and supports run-time refinement in real-world deployments.

This paper presents two main contributions. 1) We propose a real-world, open source\footnote{\url{https://github.com/UVA-BezzoRobotics-AMRLab/cbf_tracking}} CBF implementation with SAC-based $\alpha$ adaptation for online safety constraint refinement, designed to work with any low-level controller and motion planning policy. By learning the relationship between $\alpha$ and the robot's state, desired control input, and sensed obstacle configuration, the adaptation policy effectively balances safety and progress toward the goal. 2) We develop a SAC training pipeline that operates outside the OpenAI Gym framework, allowing the policy to train in high-fidelity simulations and refine itself in real-world deployments. However, we note that in our experimental trials, online adaptation was not necessary to ensure system safety while navigating to the goal.

\section{Related Work} \label{sec:rel_lit}

While motion planning is an active field of research within the robotics community, the challenge of achieving safe, agile navigation in cluttered, unknown environments remains a challenge \cite{xiao2024barn}. CBFs \cite{ames2014firstcbf,ames2017zcbf}, grounded in Nagumo's theorem on set invariance \cite{nagumo1942invariance}, have emerged as a promising approach to address safety for motion planning. In \cite{koushil2022polycbf}, a CBF-based filter for navigating cluttered environments is introduced, while \cite{li2023cbf} explores CBF formulations for source-seeking tasks. However, these methods assume convex obstacles and are validated in simulation only. \cite{pen2023logitcbf} applies logistic regression-based CBF constraints for obstacle clusters on a physical platform. However, a common drawback across these approaches is that traditional CBF formulations often lead to overly conservative behavior, such as deadlocks.

To address conservatism, \cite{dai2022learncbf} expand the safe set of control inputs by leveraging a deep differential network to learn a residual term in the CBF. While promising, the approach remains limited to simulations, and the learned CBF is a black box, offering no intuition for the learned safety constraints. Rather than learning the CBF, \cite{ames2020adaptive} introduces adaptive CBFs (aCBFs), which dynamically adjust the CBF constraints to maintain safety. However, this method remains conservative since it prevents the system from approaching the boundary of the safe set. To address this \cite{lopez2020racbf} introduce Robust aCBFs (RaCBFs), which relax the aCBF adaptation laws and allow the system to approach the safe set boundary without compromising safety. Along similar lines, and most relevant to our work, \cite{parwana2023rtcbf} propose Rate-Tunable CBFs (RT-CBFs), which update the CBF safety constraints online by adapting the class-$\mathcal{K}$ function parameters. However, these adaptive approaches have only been deployed in simulation, and real world implementations are challenging due to their reliance on auxiliary signals that require complex, unintuitive design and computation steps \cite{islam2015adaptive}. 

In regards to robust trajectory tracking techniques, \cite{fukuda2020cbftrajectory} propose a Lyapunov-based trajectory controller that integrates relaxed CBF constraints for obstacle avoidance. However, the approach still relies on a traditional, conservative CBF formulation. Beyond CBF methods, \cite{chen2020fastrack} introduce a robust tracking controller using Hamilton-Jacobi-Isaacs (HJI) reachability analysis. Due to the high computational cost of HJI, however, the approach relies on a low-fidelity planning model, compromising tracking performance.

To our knowledge, the proposed approach is the first application of the SAC algorithm to adapt CBF constraints at runtime in a general motion planning pipeline, improving robustness across multiple robotic platforms in both simulation and experiments without excessive conservatism. 

\section{Control Barrier Function Preliminaries}\label{sec:preliminaries}

Consider the following nonlinear, control-affine system:
\begin{equation}\label{eq:nonlinear_sys}
    \dot{\bm{x}} = f(\bm{x}) + g(\bm{x})\bm{u}, 
\end{equation}
where $\bm{x} \in \mathcal{X} \subset \mathbb{R}^n$ is the state, $\bm{u} \in \mathbb{R}^m$ is the control input, and $f$ and $g$ are locally Lipschitz, continuous functions defining the system dynamics.
To ensure safety, the system state must remain within a designated ``safe'' set $\mathcal{C} \subseteq \mathcal{X}$. This safe set $\mathcal{C}$ is defined as the superlevel set of a continuously differentiable function $h : \mathcal{X} \rightarrow \mathbb{R}$:
\begin{align} \label{eq:safe_set}
\mathcal{C} &= \left\{\bm{x} \in \mathcal{X} \; : \; h\left(\bm{x}\right) \geq 0\right\}  \\
\partial\mathcal{C} &= \left\{\bm{x} \in \mathcal{X} \; : \; h\left(\bm{x}\right) = 0\right\}  \\
\text{Int}\left(\mathcal{C}\right) &= \left\{\bm{x} \in \mathcal{X} \; : \; h\left(\bm{x}\right) > 0\right\} 
\end{align}
where $\partial\mathcal{C}$ and $\text{Int}\left(\mathcal{C}\right)$ denote the boundary and interior of $\mathcal{C}$ respectively. When defined in this way, the safety of the system is guaranteed so long as the system state $\bm{x}_t$ remains within $\mathcal{C}$ for all $t \geq 0$. To enforce this safety condition, $\mathcal{C}$ must be forward invariant under the system dynamics in \eqref{eq:nonlinear_sys}. More formally, the set $\mathcal{C}$ is said to be forward invariant if, $\forall \bm{x}_0 \in \mathcal{C}$, it holds that $\bm{x}_t \in \mathcal{C}$ for all $ t \geq 0$. In conjunction with \eqref{eq:safe_set}, it follows that if $h(\bm{x}) \geq 0$, then the system remains within the safe set $\mathcal{C}$, thereby ensuring safety.

In this work, we maintain the safety condition $h(\bm{x}) \geq 0$ by leveraging the widely-used Zeroing Control Barrier Function (ZCBF) \cite{ames2017zcbf}. By imposing the following ZCBF constraint, the non-negativity of $h(\cdot)$ can be guaranteed:
\begin{equation}\label{eq:cbf_const}
    \sup_{\bm{u} \in \mathcal{U}} \left[L_fh(\bm{x}) + L_gh(\bm{x})\bm{u} + \alpha_e(h(\bm{x}))\right] \geq 0,
\end{equation}
for all $\bm{x} \in \mathcal{X}$, where $\alpha_e(\cdot)$ is an extended class-$\mathcal{K}$ function, typically of the form $\alpha_e(x) = \alpha x$, with $\alpha \in \mathbb{R}$ being a user-defined parameter controlling the admissible input set. A larger $\alpha$ relaxes \eqref{eq:cbf_const}, allowing more aggressive control but reduced safety conservatism. Conversely, a smaller $\alpha$ restricts the control set, enforcing more cautious behavior. This presents a challenge when tuning $\alpha$, as it involves a difficult trade-off between safety and control performance. 

To address this, a common approach is to apply the ZCBF constraint as a filter over the control inputs $\bm{u}_\Lambda$ generated by a general low-level controller $\Lambda: \mathbb{R}^n \rightarrow \mathbb{R}^m$. This safety filtering can be formulated as a ZCBF Quadratic Program (ZCBF-QP) whose objective is to find an input $\bm{u}$ that minimally adjusts $\bm{u}_\Lambda$ to satisfy \eqref{eq:cbf_const}:
\begin{equation}\label{eq:cbf_qp}
\begin{aligned}
    & \underset{\bm{u}}{\text{argmin}}
    & & \norm{\bm{u}_\Lambda - \bm{u}}^2 \\
    & \text{subject to}
    & & L_fh(\bm{x}) + L_gh(\bm{x})\bm{u} + \alpha h(\bm{x}) \geq 0.
\end{aligned}
\end{equation}

While this formulation offers a potential solution for balancing safety and control performance, its effectiveness still relies on precise tuning of $\alpha$.

\section{Problem Formulation} \label{sec:problem}  
In this work we focus on developing a method to adapt the ZCBF safety constraint in \eqref{eq:cbf_const} to improve the {\editcolor safety} of a general low-level trajectory tracking controller $\Lambda$. Let \eqref{eq:nonlinear_sys} define the equations of motion for a mobile robotic system tasked to navigate an unknown environment. Traditionally, a receding horizon motion planning policy $\Pi$ is used to generate a time parameterized trajectory $\bm{r}(t; t_0) \in \mathbb{R}^n$ on the time interval $[t_0, t_f]$, where $t_f=t_0 + T_r$ and $T_r$ is the time horizon of the trajectory. The purpose of this trajectory is to provide a high-level path plan from the vehicle's current state $\bm{x}_t^0$ to a goal $\bm{x}_g$ while avoiding a state subset $\mathcal{X}_O(t_0)$ of obstacles currently known to the vehicle. 

While tracking $\bm{r}(\cdot)$, information about obstacles are updated such that, in general, $\mathcal{X}_O(t) \not= \mathcal{X}_O(t_0)$. Consequently, the path planner may fail to update the trajectory in response to newly sensed obstacles due to infeasible constraints or unmodeled disturbances. Additionally, since $\bm{r}(\cdot)$ is typically optimized for speed, it may cut too close to obstacles, leaving little margin for error. As a result, even if $\bm{r}(\cdot)$ is obstacle free, collisions may still occur if the low-level controller $\Lambda$ cannot faithfully track the reference trajectory due to model mismatches or input bounds. In all these cases, the responsibility of safety assurance falls to $\Lambda$. 

\textbf{Problem 1 -- Safe Tracking Control: } Given a policy $\Pi$ that generates a reference trajectory $\bm{r}(t; t_0)$ from the current state $\bm{x}_t^0$ to goal $\bm{x}_g$ while avoiding the obstacle set $\mathcal{X}_O(t_0)$:
\begin{equation}
    \bm{r}(t; t_0) = \Pi(\bm{x}_t^0, \bm{x}_g, \mathcal{X}_O(t_0)),
\end{equation}
the objective of the safe tracking control problem is to design a general low-level controller $\Lambda$ that generates an input signal $\bm{u}_t$ to track $\bm{r}(\cdot)$ while ensuring system safety. The controller $\Lambda$ produces $\bm{u}_t$ based on $\bm{x}^0_t$ and $\bm{r}(\cdot)$:
\begin{equation}
    \bm{u}_t = \Lambda(\bm{x}^0_t, \bm{r}(t;t_0)).
\end{equation}

In this work, we focus on realizations of $\Lambda$ that leverage ZCBFs, using the constraint in \eqref{eq:cbf_const} to keep the system safe.
However, tuning the $\alpha$ parameter poses a significant challenge. A constant value can be overly conservative in one region of the environment, causing deadlocks, while being too aggressive in others, compromising safety. To address this, $\alpha$ should be adapted dynamically based on the current environment and vehicle state.




\textbf{Problem 2 -- Adaptive Safe Tracking Control:} Given a ZCBF-enabled low level controller $\Lambda$, we seek to find an adaptation policy {\editcolor $\pi_\theta(\cdot | \bm{s}_t)$ that adjusts $\alpha$ in real-time. This policy should use a state embedding $\bm{s}_t$, which includes $\mathcal{X}_O(t_0)$, $\bm{x}_t$, and desired input $\bm{u}_t$}, to ensure the vehicle safely reaches $\bm{x}_g$ while avoiding $\mathcal{X}_O(t_0)$. 

In the following sections, we discuss in detail the design of our SAC-based policy $\pi_\theta$ for adaptation of the $\alpha$ parameter, and validate our approach with extensive simulation and experimental results. 

\section{Approach}\label{sec:approach}

\begin{figure}[ht!]
    \centering
    \includegraphics[width=.48\textwidth]{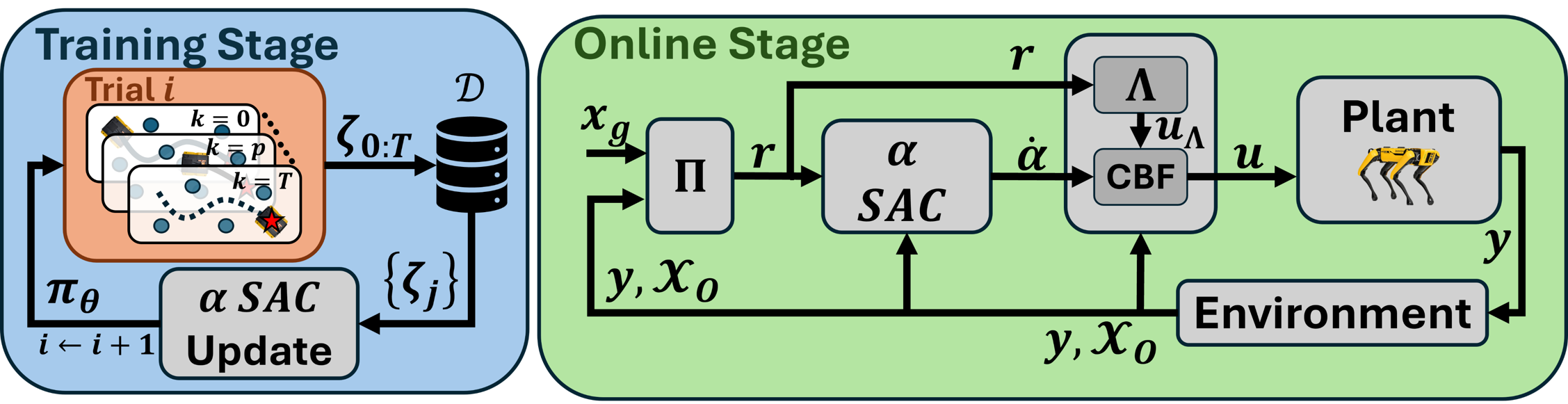}
    \caption{Block diagram for SAC training and online deployment.}
    \label{fig:approach_diagram}
\end{figure}

\begin{figure*}[ht!]
    \centering
	\subfigure[]{\includegraphics[width=0.129\textwidth]{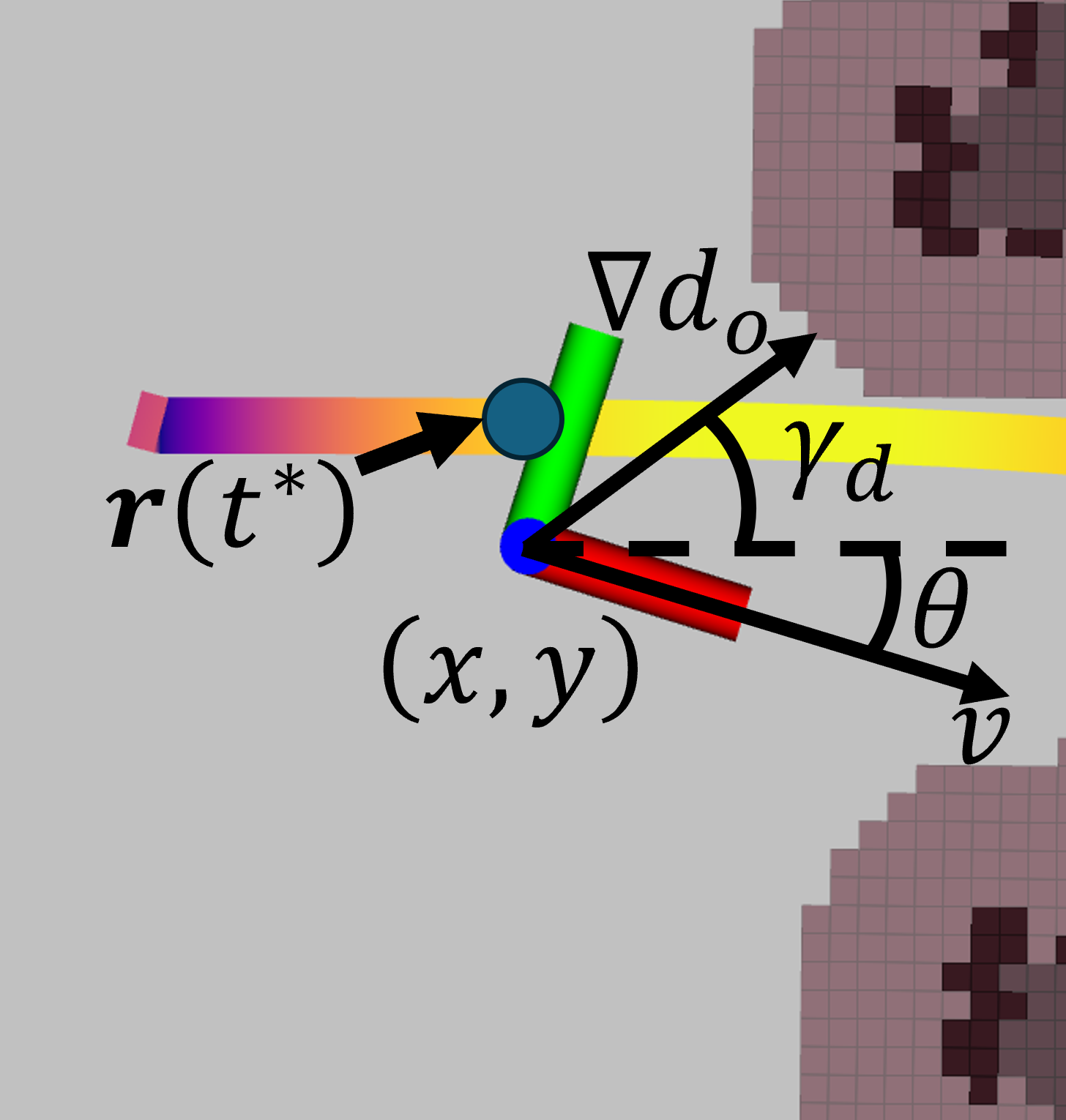}\label{fig:state_variables}}
	\subfigure[]{\includegraphics[width=0.20\textwidth]{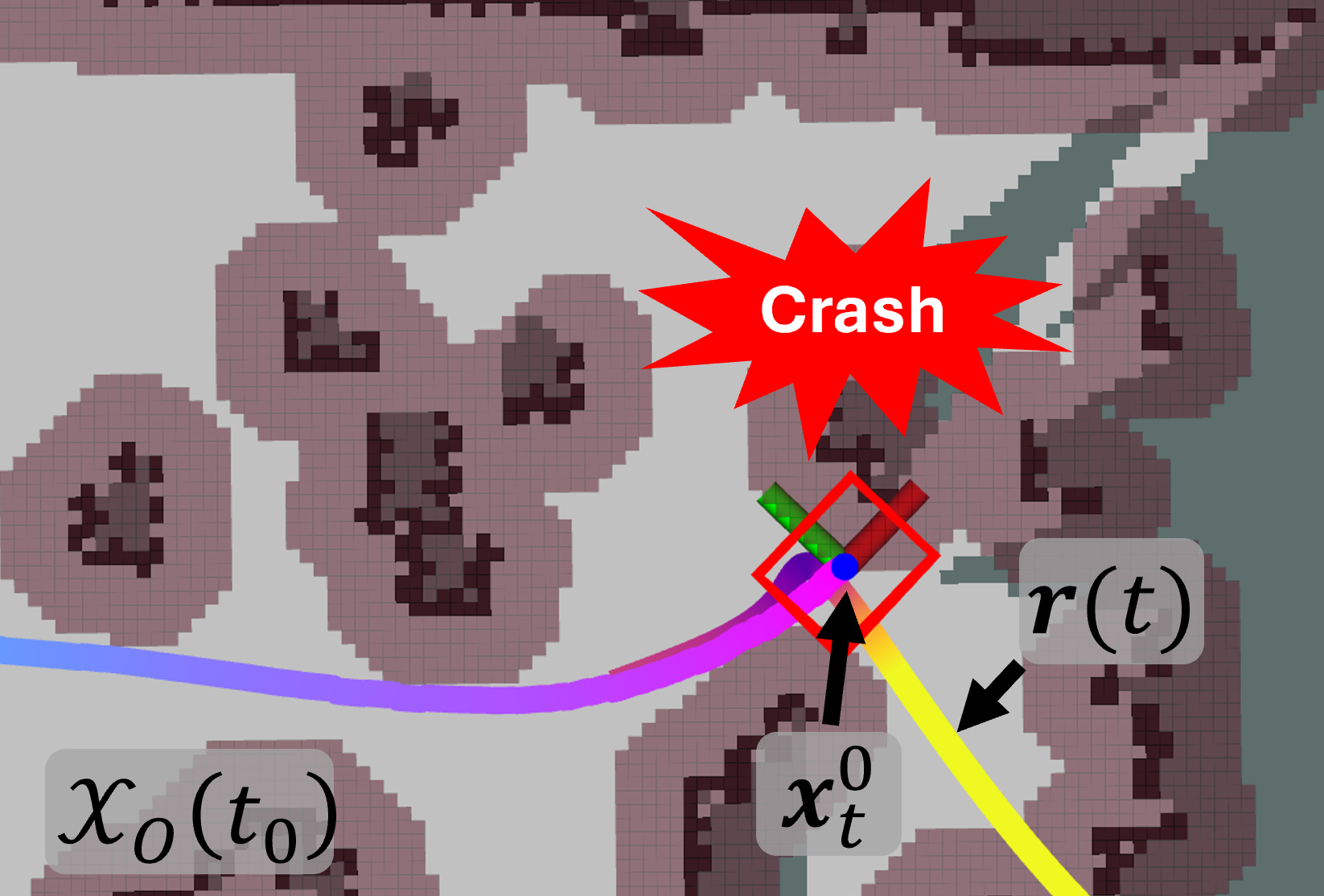}\label{fig:trajectory_gen_demo}}
	\subfigure[]{\includegraphics[width=0.193\textwidth]{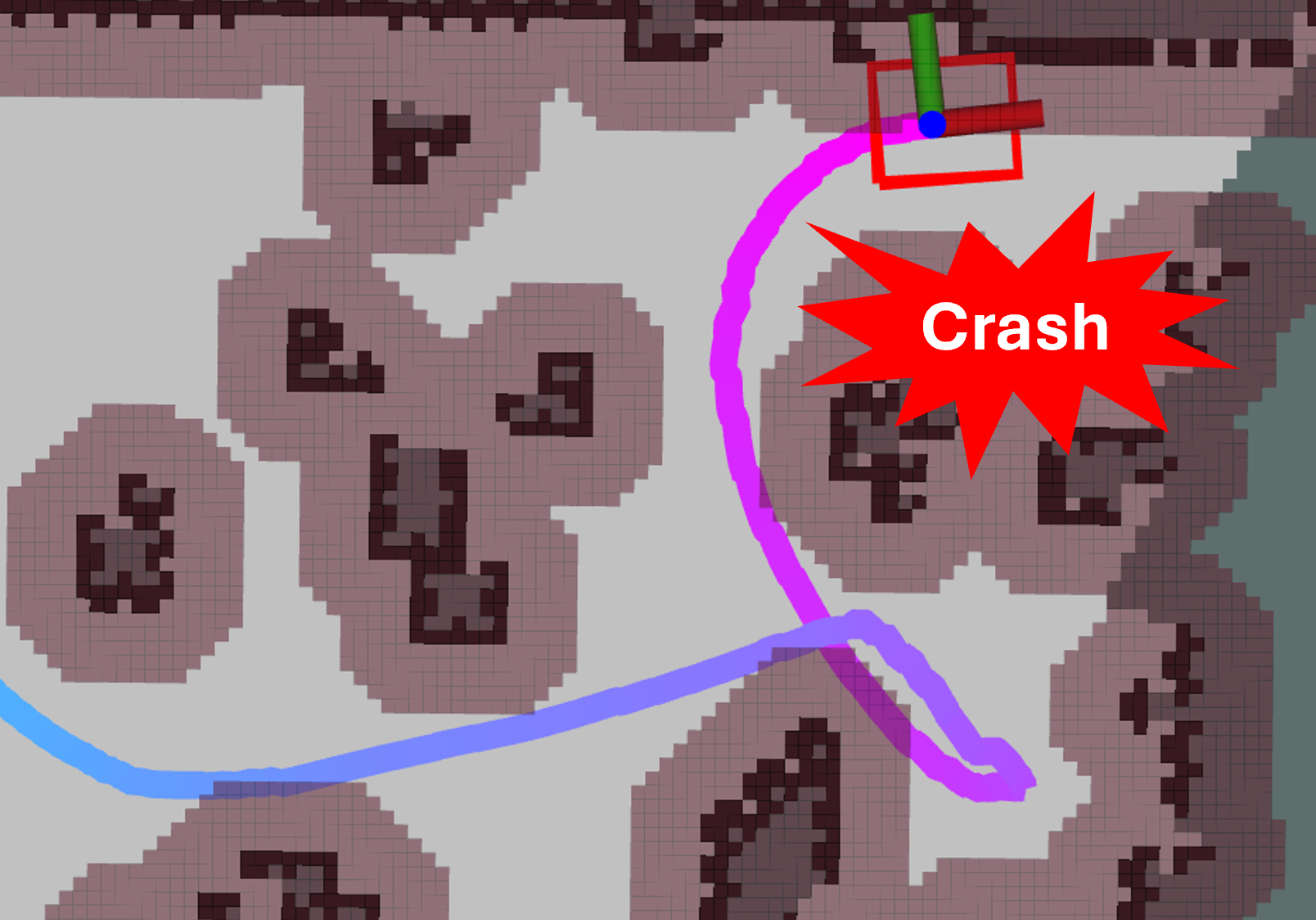}\label{fig:const_cbf_demo}}
    \subfigure[]{\includegraphics[width=0.20\textwidth]{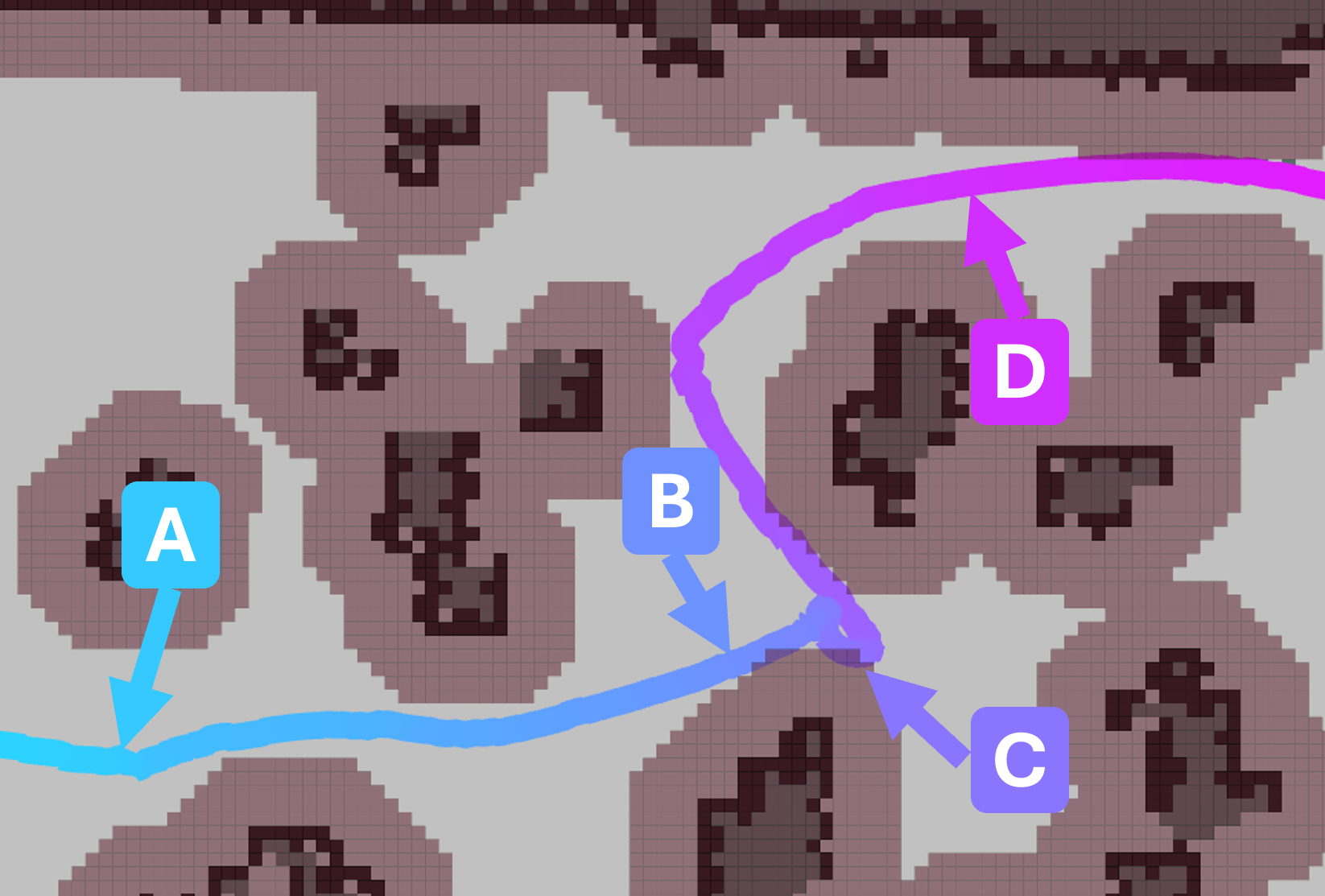}\label{fig:sac_cbf_demo}} 
    \subfigure[]{\includegraphics[width=0.222\textwidth]{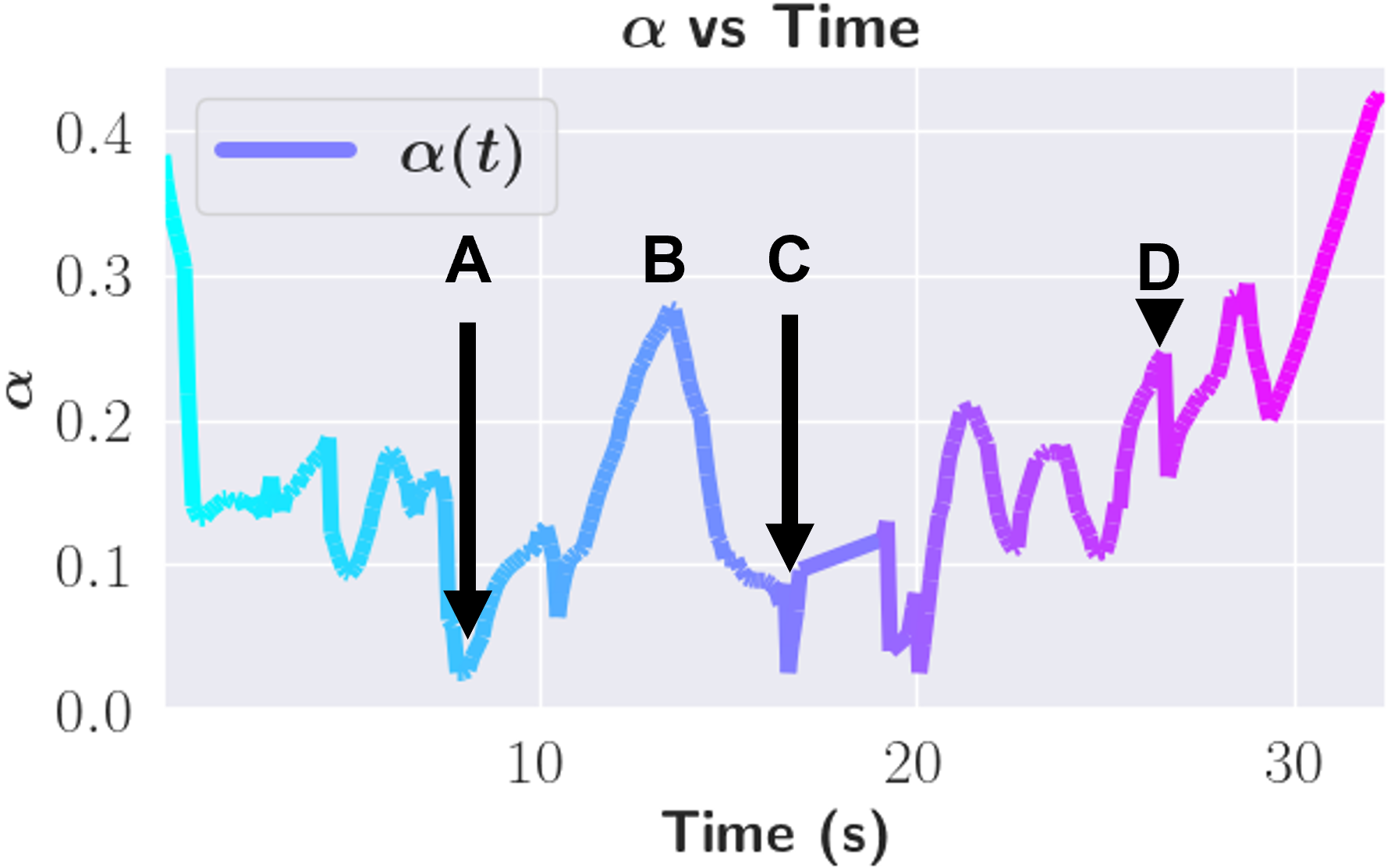}\label{fig:cbf_plot_demo}} 
    \vspace{-8pt}
    \caption{(a) Illustration of SAC state $\bm{s}_t$. (b) Base navigation pipeline crashing while tracking a generated trajectory $\bm{r}(\cdot)$. (c) Using CBF safety filter with $\alpha=.5$ still results in a collision due to infeasible constraints. (d)-(e) Full approach navigating the environment while adapting $\alpha$ as needed.}
    \vspace{-15pt}
	\label{fig:motivating_demo}
\end{figure*}

We propose a SAC-based control barrier adaptation scheme that dynamically adjusts the $\alpha$ parameter in \eqref{eq:cbf_const}, enhancing safety for trajectory tracking in cluttered and unknown environments without requiring manual, environment specific-tuning. Fig.~\ref{fig:approach_diagram} outlines our approach. Data are collected at each control time-step $k$ during simulated navigation trials as state transitions $\bm{\zeta}_k$ into a replay buffer $\mathcal{D}$, which is used to train the SAC-based $\alpha$ adaptation policy $\pi_\theta(\cdot | \bm{s}_t)$ given environmental embedding $\bm{s}_t$ (illustrated in Fig.~\ref{fig:state_variables}). For navigation, a receding horiozn motion planning policy $\Pi$ produces a $C^2$-continuous, time parameterized trajectory $\bm{r}(t;t_0) \in \mathbb{R}^2$, which is tracked by a low-level controller $\Lambda$. However, as shown in Fig.~\ref{fig:trajectory_gen_demo}, relying solely on $\Pi$ and $\Lambda$ for motion planning, without incorporating any safety scheme, can result in collisions if the low-level controller is unable to faithfully track $\bm{r}(\cdot)$. While using a ZCBF filter with a constant $\alpha$ can improve safety, selecting an appropriate $\alpha$ remains challenging. A lower $\alpha$ risks deadlocks, while a larger $\alpha$, as in Fig.~\ref{fig:const_cbf_demo}, may be too lenient, allowing the vehicle to get closer to obstacles and require controls above actuation limits to avoid collisions.

To address these limitations, we propose a ZCBF scheme with a time-varying $\alpha(t)$ that filters the control inputs of $\Lambda$. This allows \eqref{eq:cbf_const} to dynamically adapt at each control cycle through $\pi_\theta$, based on the current vehicle state $\bm{x}^0_t$, desired input $\bm{u}^0_t$, and sensed obstacle configuration $\mathcal{X}_O(t_0)$. When using this adaptation scheme, as shown in Figs.~\ref{fig:sac_cbf_demo} and \ref{fig:cbf_plot_demo}, $\alpha(t)$ decreases when the vehicle approaches obstacles (points A and C), increasing conservatism and pushing the vehicle away. Conversely, $\alpha(t)$ increases when entering more open spaces (point B) or when a low $\alpha(t)$ would prevent passing through narrow openings (point D). In the following sections, we describe our $\alpha$ adaptation framework in detail, starting with the formulation of the ZCBF used in this work.

\subsection{ZCBF Formulation}
{\editcolor While the proposed SAC-based $\alpha$ adaptation framework is designed for a general system $\dot{\bm{x}}$, we focus on a second-order unicycle model since it is applicable to a wide range of mobile platforms.} The system state $\bm{x}$ consists of $\text{SE}(2)$ pose $[x,y,\theta] \in \mathcal{X} \subset \mathbb{R}^2 \times [-\pi, \pi]$ and velocity $v \in \mathbb{R}$, and its dynamics are given by:
\begin{equation}\label{eq:unicycle_dynamics}
    \dot{\bm{x}} =
    \begin{bmatrix}
        \dot{x} \\
        \dot{y} \\
        \dot{\theta} \\
        \dot{v}
    \end{bmatrix}
    =
    \begin{bmatrix}
        v \cos(\theta) \\
        v \sin(\theta) \\
        0 \\
        0
    \end{bmatrix} + 
    \begin{bmatrix}
        0 & 0 \\
        0 & 0 \\
        0 & 1 \\
        1 & 0
    \end{bmatrix}
    \begin{bmatrix}
        a \\
        \omega
    \end{bmatrix},
\end{equation} 
where $\bm{u} = \left[a \enspace \omega\right]^T$ is the input vector consisting of linear acceleration $a$ and angular velocity $\omega$. 

To ensure system safety, we define the safe set $\mathcal{C}$ as the set of states where the distance $d_o(x,y)$ between the vehicle's center $(x,y)$ and nearest obstacle is larger than the vehicle's minimum circumscribing radius $r_v$:
\begin{equation}
\mathcal{C} = \{\bm{x} \in \mathcal{X} \; | \; d_o(x, y) - r_v > 0\}.
\end{equation}

Using this definition of $\mathcal{C}$, the ZCBF is formulated as proposed in \cite{li2023cbf,ames2017zcbf}:
\begin{equation}\label{eq:h_func_def}
    h(\bm{x}) = [d_o(x,y)-r_v]\exp{\left\{\nabla d_o(x, y) \cdot \bm{e}_{\theta} - vd_1\right\}},
\end{equation}
where $\bm{e}_{\theta} = [\cos \theta \; \sin \theta]$ is the unit orientation vector of the vehicle, $d_1 \in \mathbb{R}^+$ is a user-defined parameter ensuring $vd_1\in (0,1)$, and $\nabla d_o(x,y) = [\frac{\partial d_o}{\partial x} \; \frac{\partial d_o}{\partial y}]$ is the spatial gradient of $d_o(x,y)$. In this work $d_o(\cdot)$ and $\nabla d_o(\cdot)$ are computed using a Euclidean Signed Distance Field (ESDF).




\subsection{Dynamic $\alpha$ Adjustment}\label{subsec:learning_cbf_sac}
While the ZCBF formulation introduced in the previous section generally enhances safety as the vehicle navigates towards its final goal $\bm{x}_g$, using a constant $\alpha$ poses two challenges. First, tuning $\alpha$ to achieve safety while preventing deadlocks is difficult due to the varying obstacle distribution within an environment, (see Fig.~\ref{fig:const_cbf_demo}). Second, even if an appropriate $\alpha$ is found, it won't be suitable for all possible environments. These challenges create an over-constrained problem, making runtime adaptation of $\alpha$ necessary. To address this, the $\alpha$ parameter in \eqref{eq:cbf_const} can be adapted dynamically within a user-defined interval $\alpha(t) \in [\alpha^-, \alpha^+]$ (detailed in Sec.~\ref{sec:simulation}), allowing the safe set of inputs to expand or contract as the vehicle navigates its environment. 

To adapt $\alpha(t)$, we learn a policy $a_t\sim \pi_\theta(\cdot | \bm{s}_t)$, where the input $\bm{s}_t \in \mathbb{R}^{N_s}$ is an environmental state embedding and the action $a_t$ is the derivative $\dot{\alpha}(t)$. The adaptation of $\alpha(t)$ can then be governed via the following update equation:
\begin{equation}\label{eq:alpha_update}
    \alpha(t) = \alpha(0) + \int_0^t \dot{\alpha}(\tau) \, d\tau,
\end{equation}
where $\dot{\alpha}(\tau) \sim \pi_\theta(\cdot | \bm{s}_\tau)$ and $\alpha(0)$ is a user-defined initial value. In practice, we observed that $\alpha(t)$ can be updated quickly enough that the choice of $\alpha(0)$ does not significantly impact performance. Therefore, we set $\alpha(0)=\alpha^+$ to avoid conservatism from the outset. 
To implement $\pi_\theta(\cdot | \bm{s}_t)$, we use the Soft Actor-Critic (SAC) RL algorithm as it effectively handles continuous action spaces and offers efficient sampling during training. Additionally, its stochastic nature is well-suited to handle the uncertainty and noise prevalent in real-world robotic applications.
\subsection{SAC for $\alpha$ Adaptation} The SAC is a state-of-the-art, off-policy algorithm for real-world robotic reinforcement learning \cite{haarnoja2019sac}. A key feature is its objective function, which seeks a stochastic policy $\pi_\theta$ that maximizes expected return while also maximizing entropy:
\begin{equation}\label{eq:sac_return}
    J_\pi = \mathop{\mathbb{E}}_{\bm{\tau} \sim \pi}\left[\sum_{t=0}^\infty \gamma^t \Big( R(\bm{s}_t, a_t, \bm{s}_{t+1}) + \beta H(\pi_\theta(\cdot | \bm{s}_t)\Big)\right],
\end{equation}
where $\gamma \in (0, 1]$ is a discount factor on future rewards $R(\cdot)$, $\bm{\tau} = \left(\bm{s}_0, a_0, \bm{s}_1, a_1, \dots \right)$ is a sequence of states $\bm{s}_t$ and actions $a_t$ in the environment, $H(P)=\mathbb{E}_{x \sim P}[-\log P(x)]$ is the entropy term, and $\beta$ is a fixed trade-off coefficient that is automatically adjusted during training \cite{haarnoja2019sac}. This dual objective encourages a balance between exploration -- incentivizing the vehicle to explore new states -- and exploitation -- maximizing the expected return. As a result, the objective prevents $\pi_\theta$ from converging to sub-optimal local minima, while also improving sample efficiency and policy stability.

As the vehicle navigates through the environment, it adds a collection of state transition tuples $\bm{\zeta}_t = (\bm{s}_{t}, a_{t}, r_{t}, \bm{s}_{t+1}, d_t)$ to a replay buffer $\mathcal{D}$ of maximum size $N_d$, where $r_{t}$ is the transition reward and $d_t$ indicates whether $\bm{s}_{t+1}$ is a terminal state (i.e., collision or reach final goal). During training, the SAC algorithm learns three deep neural networks: the (actor) policy $\pi_\theta$ along with two (critic) Q-functions $Q_{\phi_1}$ and $Q_{\phi_2}$, parameterized by $\theta$, $\phi_1$, and $\phi_2$ respectively. To update $\theta$ and $\phi_i$, mini-batches are sampled from $\mathcal{D}$ to perform stochastic gradient descent, minimizing the following loss functions:
\begin{align}\label{eq:sac_loss_functions}
    L(\theta, \mathcal{D}) &= \mathop{\mathbb{E}}_{\substack{\bm{s_t}\sim\mathcal{D} \\ a_t\sim\pi_\theta}} \left[-\min_{j=1,2} Q_{\phi_j}(\bm{s}_t,a_t) + \beta\log\pi_\theta(a_t|\bm{s}_t) \right], \nonumber \\
    L(\phi_i, \mathcal{D}) &= \mathop{\mathbb{E}}_{\bm{\zeta}_t \sim \mathcal{D}} \left[\left( Q_{\phi_i}(\bm{s}_t,a_t)-y(r_t,\bm{s}_{t+1},d_t) \right)^2 \right].
\end{align}

Here $y(\cdot)$ is the target for the critic functions and is computed from the received reward and target value of the next state-action pair, incorporating entropy regularization to encourage exploration \cite{haarnoja2019sac}. Using these loss functions, the SAC framework refines the stochastic policy $\pi_\theta(\cdot | \bm{s}_t)$, which dynamically adjusts $\alpha$ as the vehicle progresses toward its goal $\bm{x}_g$. However, the performance of the policy relies on well-crafted state and reward design.

\subsubsection{State and Reward Design:} To ensure the SAC policy adapts $\alpha$
while maintaining the feasibility of \eqref{eq:cbf_const}, the state representation $\bm{s}_t$ must include all relevant system and environment variables necessary for computing $h(\cdot)$ in \eqref{eq:h_func_def}, along with the value of $h(\cdot)$ itself. Additionally, we include a term $\rho$ to reward progress along the trajectory to $\bm{x}_g$.

To define $\rho$, we first determine $t^*$, the time associated with the closest point along the trajectory $\bm{r}(\cdot)$ to the vehicle's current $xy$-position $(x_t^0, y_t^0)$:
\begin{equation}
    t^* = \arg\min_{t\in [t_0, t_f]} \norm{\bm{r}(t) - \begin{bmatrix} x^0_t \\ y^0_t \end{bmatrix} }.
\end{equation}
The progress term $\rho$ is then given by the ratio of $t^*$ to the total trajectory duration $T_r=t_f-t_0$: 
\begin{equation}
    \rho = \frac{t^*}{T_r}.
\end{equation}
Thus, the full state vector $\bm{s}_t$ is defined below (see Fig.~\ref{fig:state_variables} for illustration). For brevity, we omit the time-dependence:
\begin{equation}
\bm{s}=\left[\theta \enspace v \enspace a \enspace \omega \enspace d_o(x,y) \enspace \gamma_d(x,y) \enspace \rho \enspace \alpha \enspace h(\bm{x}) \right]^T, 
\end{equation}
where $\gamma_d(\cdot)$ denotes the heading to the closest obstacle, and is defined as $\gamma_d(x,y)=\atantwo(\frac{\partial d_o}{\partial y}, \frac{\partial d_o}{\partial x})$. 


With the state defined, the reward function $R(\bm{s}_{t}, a_{t}, \bm{s}_{t+1})$ is constructed to prevent deadlocks and promote safety by heavily penalizing violations of the ZCBF constraint in \eqref{eq:cbf_const}:
\begin{equation}
    R(\cdot) = \mu_d d_o + \mu_\rho \rho - \mu_r \dot{\alpha}^2 - \mu_b b(\alpha) - \mu_h \psi(h),
\end{equation}
where $\mu_d, \mu_\rho, \mu_r, \mu_b, \mu_h \in \mathbb{R}^+$ are weighting parameters for each component of the reward function. 

Breaking down the reward function: the $d_o$ term encourages the vehicle to maintain a safe distance away from obstacles and imposes a large negative reward in the event of a collision. While this term encourages the vehicle to avoid obstacles, relying on it alone can lead to situations where the vehicle receives positive rewards even if it becomes deadlocked. To address this issue, the $\rho$ term rewards progress along the current trajectory $\bm{r}(t)$ towards the final goal.

To promote stable adaptation, the $\dot{\alpha}$ term penalizes excessive changes to $\alpha$. The function $b(\alpha)$ imposes a penalty when $\alpha(t)$ falls outside the bounds $[\alpha^-, \alpha^+]$, ensuring $\pi_\theta$ learns the appropriate limits. Finally, $\psi(h)$ penalizes actions that would violate \eqref{eq:cbf_const}, encouraging $\pi_\theta$ to keep $h(\cdot)$ non-negative and thereby keep the system safe.


\subsubsection{SAC Training:}\label{sec:sac_training} For training, we deploy a navigation stack with planning policy $\Pi$ and low level controller $\Lambda$ in a set of simulation environments. Using the BARN dataset, \cite{perille2020barn}, 
we generate $40$ cluttered Gazebo worlds for training, requiring the vehicle to navigate each $5$ times before moving to the next.
At each control cycle $k$ during navigation, state transition tuples $\bm{\zeta}_k$ are recorded into an SQLite3 database, serving as our replay buffer $\mathcal{D}$. Upon reaching a terminal state (i.e., collision or reached goal), the simulation ends and $\pi_\theta(\cdot | \bm{s}_t)$ is updated by uniformly sampling mini-batches $\{\bm{\zeta}_i\}_{i=1}^{N_d} \subset \mathcal{D}$ to perform gradient descent on \eqref{eq:sac_loss_functions}.

We have implemented the data collection and training pipeline in this way to avoid the limitations of traditional OpenAI Gym environments, which are often challenging to apply in physical robot deployments. Our pipeline is designed to function in both simulation for training and in the real world for online refinement. In real-world scenarios, instead of updating the policy after each navigation task, periodic retraining can be performed asynchronously after a number $N_r$ of new transitions are added to $\mathcal{D}$. Despite supporting online policy refinement, our physical experiments demonstrated that a pre-trained policy was sufficient for all real-world trials conducted.

\section{Simulations}\label{sec:simulation}
Simulations were performed to train the SAC policy, $\pi_\theta(\cdot | \bm{s}_t)$, and to validate the proposed approach. All simulations ran in Gazebo on Ubuntu 20.04 using ROS Noetic. The robot used was a Clearpath Robotics Jackal UGV equipped with a $270^\circ$ 2D LiDAR sensor. For training, data were collected into a replay buffer $\mathcal{D}$ for updating the actor and critic models $\pi_\theta, Q_{\phi_1}, Q_{\phi_2}$ as described in Sec.~\ref{sec:sac_training}.

We evaluated the performance of our approach on two different realizations of $\Lambda$: {\editcolor a Model Predictive Controller (MPC) \cite{fossen2015mpc} and a Proportional-Derivative (PD) \cite{gray2006geometry} controller}. For each, we tested three configurations: no ZCBF, a constant $\alpha$ for the ZCBF constraint in \eqref{eq:cbf_const}, and our full approach with SAC-based adaptation. The MPC, written in C++ using the Sequential Least SQuares Programming (SLSQP) \cite{kraft1994slsqp} algorithm from NLOPT \cite{johnson2007nlopt}, was tested up to $15$Hz, but ran at $10$Hz with a prediction horizon of $N=15$. The ZCBF-QP for the PD controller, also implemented in NLOPT, filters inputs in under $1$ms, while the PD controller itself ran at $10$Hz.
For testing, we ran $5$ trials for each of the $50$ generated BARN testing worlds, including a baseline Dynamic Window Approach (DWA) \cite{fox1997dwa} for comparison. We note that we did not compare against other CBF-based methods in the literature, as their limiting assumptions make real-world deployment challenging; instead, the constant $\alpha$ case serves as a representative baseline for these approaches.
\subsection{Case Study 1: Model Predictive Controller}
In this case study, we implement our $\alpha$ adaptation framework on top of {\editcolor a tracking MPC described \cite{fossen2015mpc} for the system described in \eqref{eq:unicycle_dynamics}. }
The trajectory $\bm{r}(t;t_0)=\left[x_r(t) \enspace y_r(t)\right]^T$ is generated in a receding horizon fashion using an augmented version of the FASTER solver \cite{tordesillas2022faster}, as detailed in our prior work \cite{mohammad2024planner}. The cost function for the MPC is:
\begin{align}
    J = &  \sum_{k=1}^{N-1} \left[\norm{\Delta\bm{x}_k}^2_{\bm{Q}} + \norm{\Delta \bm{u}_k}^2_{\bm{P}}\right] + \norm{\Delta\bm{x}_N}^2_{\bm{M}}
\end{align}
where $\Delta\bm{x}_k=\bm{x}_k-\bm{x}_r(k\Delta t)$ and $\Delta\bm{u}_k=\bm{u}_k-\bm{u}_r(k\Delta t)$ are the state and input error with respect to $\bm{r}(\cdot)$, and $\Delta t$ is the sampling time. $\bm{Q}$, {\editcolor $\bm{P}$}, and $\bm{M}$ are weighting matrices for state, input, and final state respectively, $\bm{x}_k$ is the $k$th state in the predictive horizon, $\bm{u}_k$ is the $k$th control input, and $\bm{u}_r(\cdot)=\left[\norm{\ddot{\bm{r}}(\cdot)} \enspace \omega_r(\cdot) \right]$ is the reference control input, where $\omega_r$ is the reference angular velocity, as calculated in \cite{gray2006geometry}.


To accurately track $\bm{r}(\cdot)$, we augment \eqref{eq:unicycle_dynamics} with two state terms: lateral error $y_e(t)$ and heading error $\theta_e(t)$,
\begin{equation}\label{eq:track_err}
    y_e(t) =  \bm{e}^\perp_p(t)\bm{R}(\theta_r(t)), \enspace \theta_e = \theta(t) - \theta_{r}(t).
\end{equation}
Here $\theta_r(t)=\atantwo\left(\dot{y}_r(t),\dot{x}_r(t)\right)$, $\bm{e}^\perp_p(t)$ is the position error normal and $\bm{R}(\cdot) \in \mathbb{R}^{2\times2}$ is the rotation matrix. With the state and objective defined, the final Optimal Control Problem (OCP) incorporating the ZCBF constraint is formulated:
\begin{equation}\label{eq:ocp_no_cbf}
\begin{aligned}
    & \underset{\bm{x}, \bm{u}}{\text{argmin}}
    & & J(\bm{x}, \bm{u}, \bm{r}) \\
    & \text{subject to}
    & & \bm{x}_0 = \bm{x}_t^0 \\
    & & & \bm{x}_{k+1} = f(\bm{x}_k) + g(\bm{x}_k)\bm{u}_k \\
    & & & L_fh(\bm{x}_k) + L_gh(\bm{x}_k)\bm{u} _k+ \alpha h(\bm{x}_k)\geq 0 \\
    & & & \bm{x}_k \in \mathcal{X}, \enspace \bm{u}_k \in \mathcal{U}
\end{aligned}
\end{equation}

Incorporating the ZCBF within the OCP ensures proactive safety over the prediction horizon as the system approaches its goal $\bm{x}_g$. To reduce computation costs, $\alpha(t)$ is updated at the start of each control cycle using \eqref{eq:alpha_update} and assumed constant over the horizon. Additionally, to keep $\alpha(t)$ from growing unbounded, we constrain it to $[.025, .5]$ based on empirical results: $\alpha=.025$ is overly conservative, while $\alpha=.5$ minimally alters the input while still enhancing safety.

\begin{figure}[ht!]
    \centering
	\subfigure[]{\includegraphics[width=0.114\textwidth]{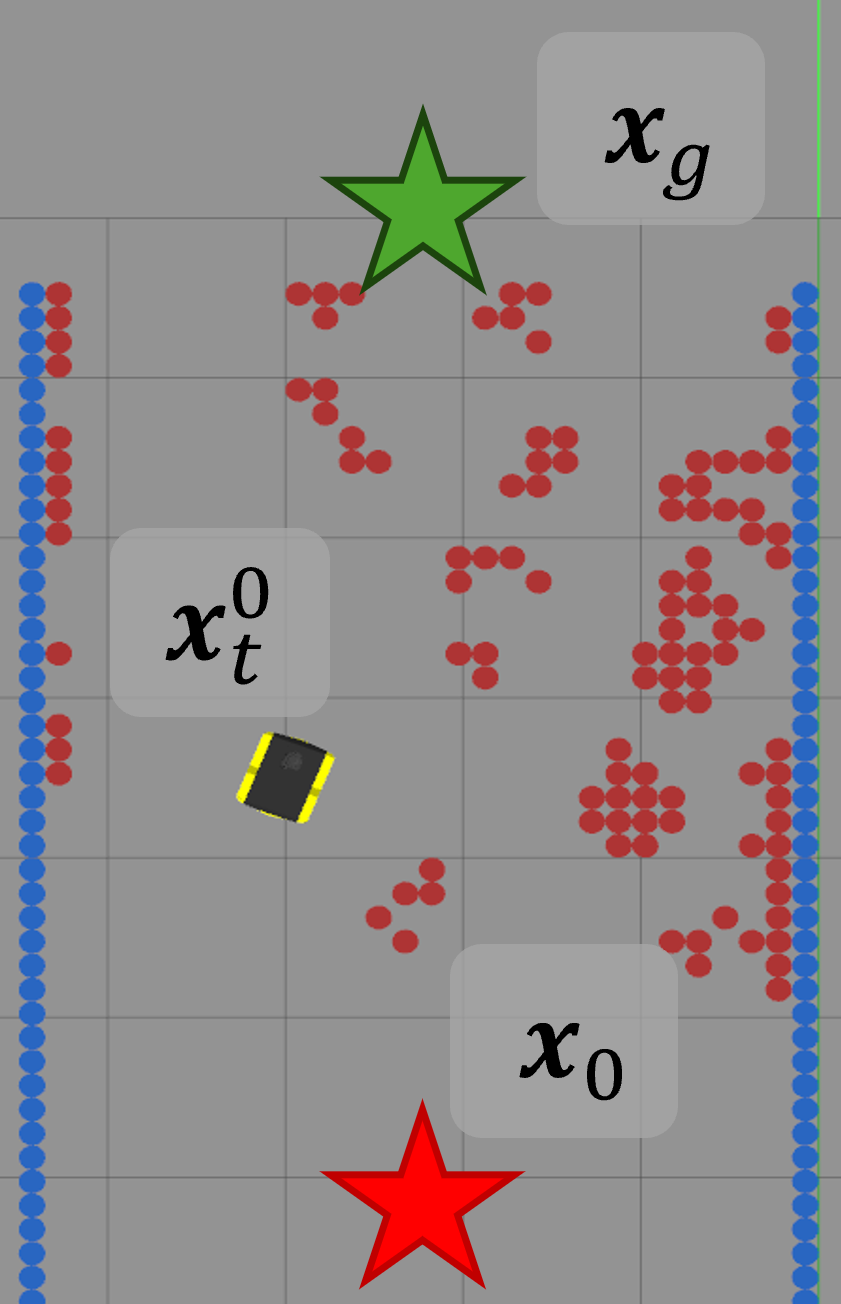}\label{fig:barn_world_5}}
	\subfigure[]{\includegraphics[width=0.114\textwidth]{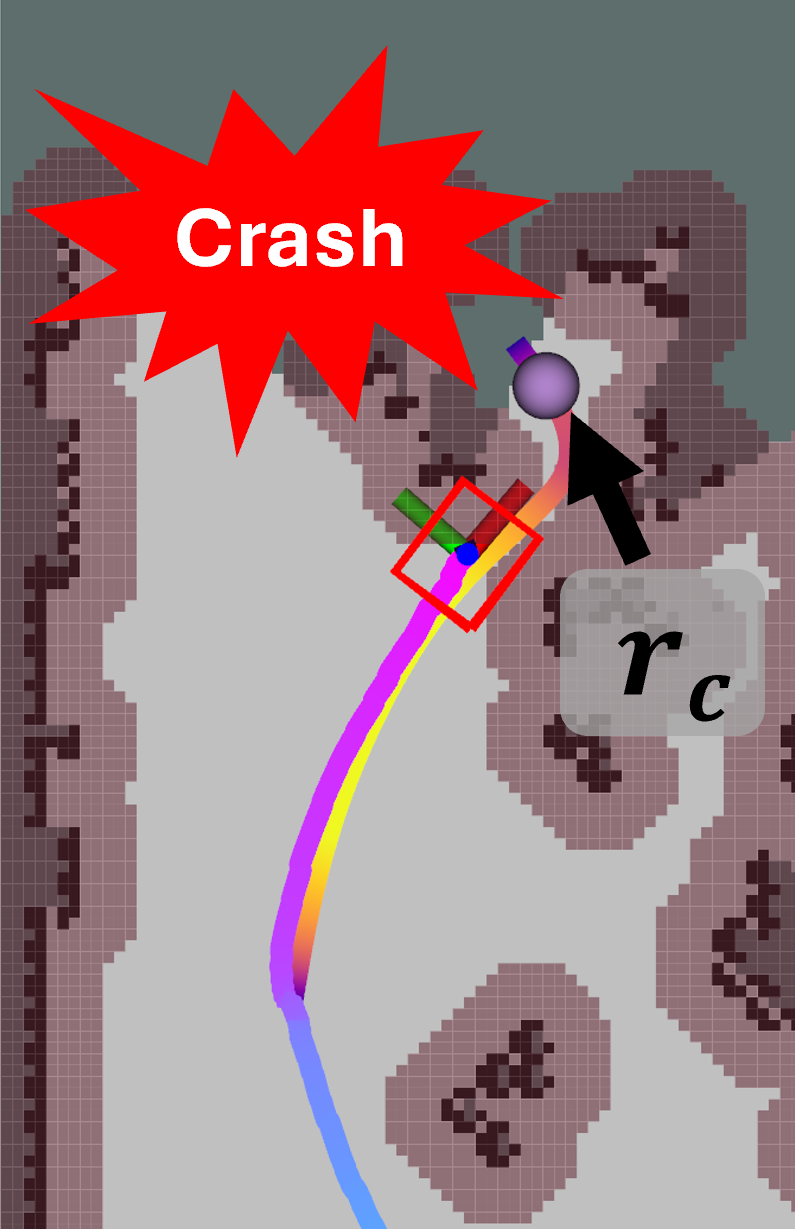}\label{fig:const_alpha_world_5}}
    \subfigure[]{\includegraphics[width=0.115\textwidth]{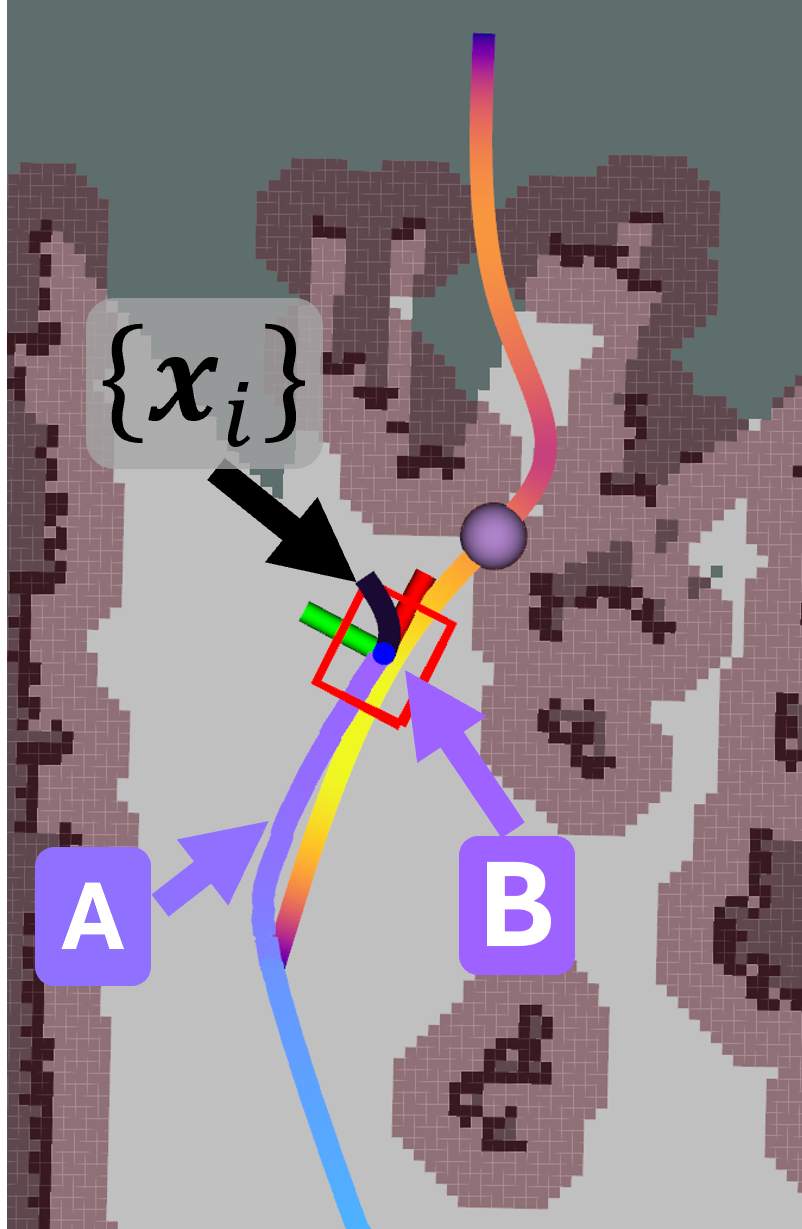}\label{fig:cbf_sac_sim_1}} 
    \subfigure[]{\includegraphics[width=0.115\textwidth]{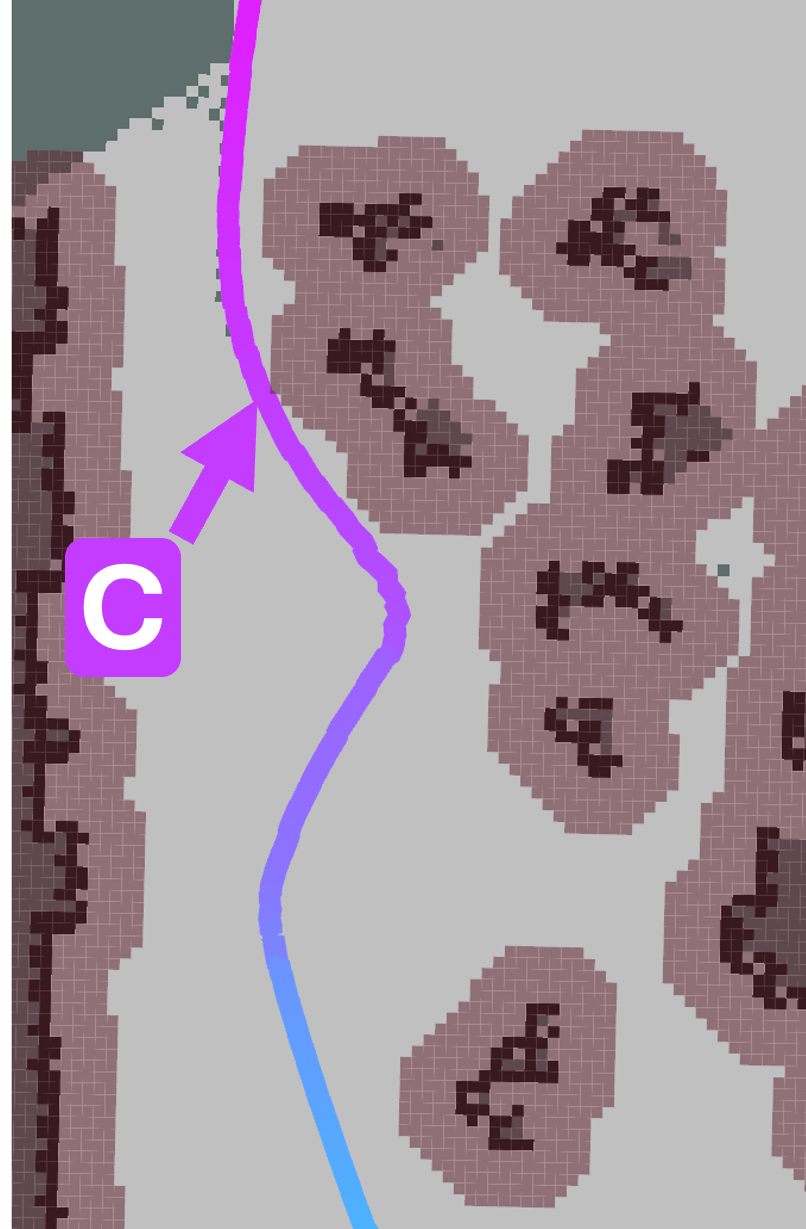}\label{fig:cbf_sac_sim_2}} \\
    \vspace{-5pt}
    \subfigure[]{\includegraphics[width=0.23\textwidth]{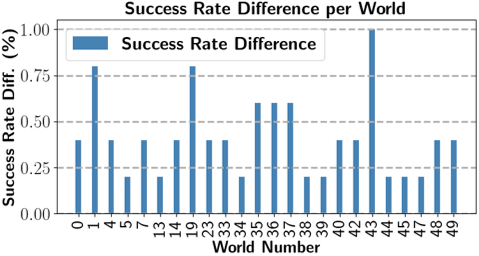}\label{fig:success_diff_plot}}
    \subfigure[]{\includegraphics[width=0.23\textwidth]{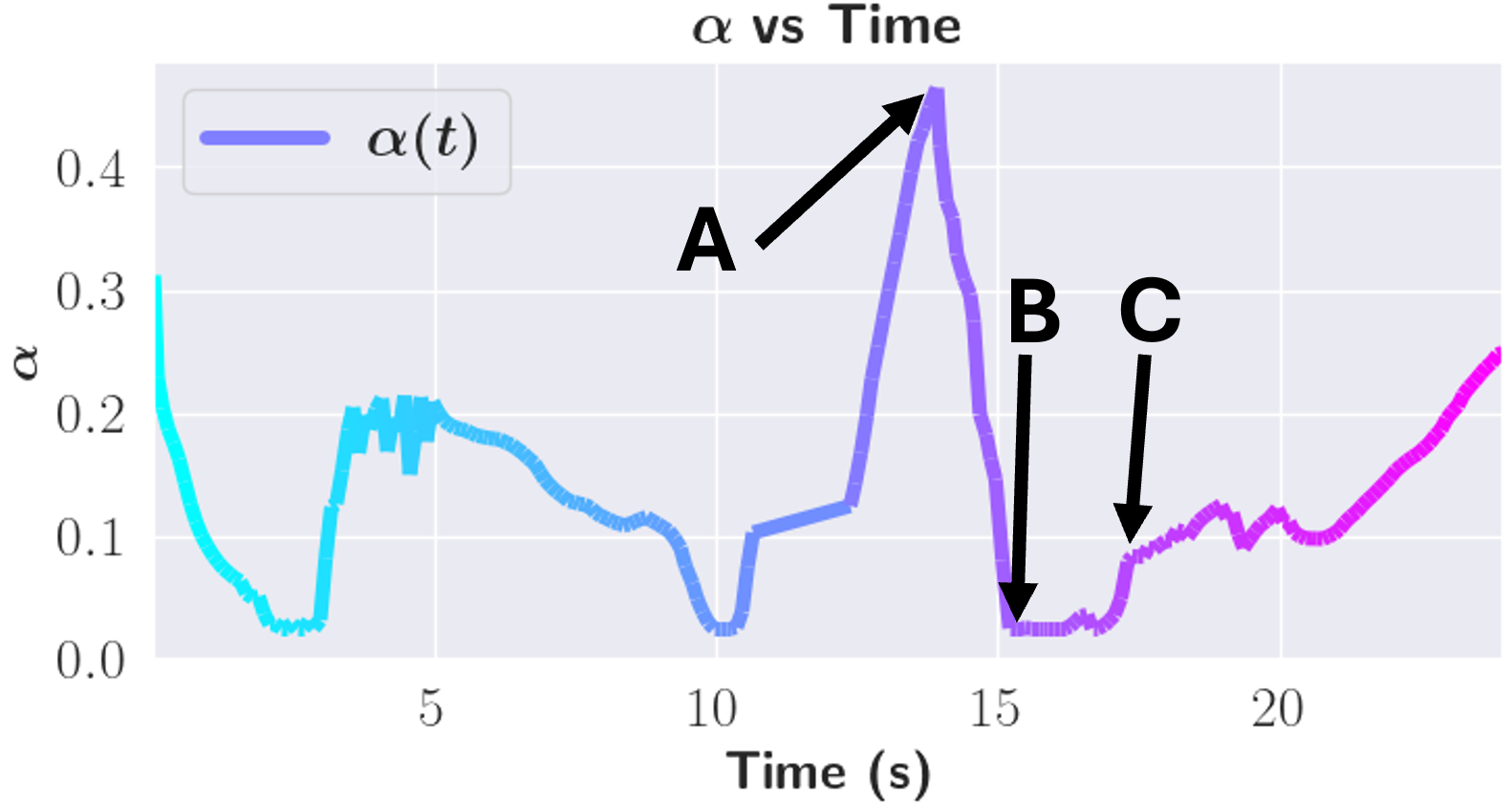}\label{fig:cbf_sac_plot_world5}} 
    \vspace{-5pt}
    \caption{Simulation results for MPC. (a) Test world 5 in Gazebo. (b) Crash with fixed $\alpha=.5$. (c), (d) Success with the full approach. (e) Success rate difference between full approach and $\alpha=.5$. (f) $\alpha(t)$ plot for full approach.}
    \vspace{-10pt}
	\label{fig:simulation_results}
\end{figure}

As shown in Table~\ref{table:sim_results}, using a constant $\alpha=.5$ in the MPC case improves the baseline MPC controller without the ZCBF by $7$\%. However, this improvement is not as pronounced as the $27$\% success rate improvement with our full SAC-based approach. Fig.~\ref{fig:success_diff_plot} illustrates the success rate differences between our full approach and the constant $\alpha=.5$ case, highlighting only those worlds where a non-zero difference was observed. Note that all of the differences are $>0$, demonstrating that our SAC-based approach consistently performs as well or better than when $\alpha$ is fixed.
\begin{table}[h!]
    \centering
    \renewcommand{\arraystretch}{1.3} 
    \setlength{\tabcolsep}{20pt} 
    \caption{Simulated Trials Success Rate Comparison}
    \vspace{-5pt}
    \begin{tabular}{|c|c|c|}
        \hline
        \textbf{Baseline} & \multicolumn{2}{c|}{\textbf{Success Rate}} \\
        \hline
        \parbox[c]{3cm}{\centering DWA} & \multicolumn{2}{c|}{$0.59$} \\
        \hline
        \textbf{ZCBF Implementation} & \textbf{MPC} & \textbf{PD} \\
        \hline
        \parbox[c]{3cm}{\centering No ZCBF} &  $0.71$ & $0.65$ \\
        \hline
        \parbox[c]{3cm}{\centering Constant $\alpha$} & $0.78$ &  $0.68$ \\
        \hline
        \parbox[c]{3cm}{\centering SAC-based} & $\bm{0.98}$ & $0.72$ \\
        \hline
    \end{tabular}
    \vspace{-5pt}
    \label{table:sim_results}
\end{table}

To illustrate how our SAC-based framework outperforms the constant $\alpha=.5$ approach, consider world $5$ as depicted in Fig.~\ref{fig:barn_world_5}, accompanied by an example of a navigation failure in the constant $\alpha$ case (Fig.~\ref{fig:const_alpha_world_5}) and success with the proposed approach (Figs.~\ref{fig:cbf_sac_sim_1} and \ref{fig:cbf_sac_sim_2}). In Fig.~\ref{fig:const_alpha_world_5}, the reference trajectory $\bm{r}(\cdot)$ leads the vehicle through a tightly cluttered region. Although the vehicle can physically fit, the narrow channel increases the likelihood of a crash. Here, the $\alpha=.5$ MPC identifies that the current reference state $\bm{r}_c$ is unsafe, and attempts to steer the vehicle away. However, since the ZCBF is not sufficiently conservative, the vehicle approaches the obstacles too closely, and is unable to avoid a collision while turning away.

Conversely, using our SAC-based approach (Figs.~\ref{fig:cbf_sac_sim_1} and \ref{fig:cbf_sac_sim_2}), $\alpha(t)$ starts decreasing at point A as the vehicle approaches the narrow entryway (Fig.~\ref{fig:cbf_sac_plot_world5}). This reduction in $\alpha(t)$ prompts the vehicle to steer away from the narrow corridor, as shown by the MPC horizon $\{\bm{x}_i\}$ (point B). As the vehicle continues through the left-side corridor, $\alpha(t)$ increases, enabling the vehicle to progress towards $\bm{x}_g$ without the ZCBF constraint unduly obstructing its path (point C).

\subsection{Case Study 2: PD Controller}
In the second case study, a PD controller for the system in \eqref{eq:unicycle_dynamics} is implemented based on $\bm{e}_p^\perp$, $y_e$, and $\theta_e$ \cite{gray2006geometry}. To incorporate the ZCBF constraint, we feed the desired control input $\bm{u}_\Lambda=[a, \omega]$ to the ZCBF-QP in \eqref{eq:cbf_qp}. For the constant $\alpha$ case, we use $\alpha=4$, which was found to be the highest performing baseline. Following the same logic as with the MPC, we restrict the bounds for $\alpha$ as $[1.5,8.0]$.


Table~\ref{table:sim_results} shows that the SAC adaptation policy also improves the performance of the baseline PD controllers, reinforcing that our approach is agnostic to the low-level controller used. However, the improvement is less pronounced compared to the MPC case ($+7$\% improvement vs +$27$\%). We believe this is due to the myopic nature of the ZCBF-QP and PD controller. While the MPC evaluates safety across the entire predicted horizon $\{\bm{x}_i\}$, the ZCBF-QP only considers safety in regards to the closest obstacle at the current time step. This also aligns with the general observation that MPC outperforms PD control due to its predictive and proactive nature.

\section{Physical Experiments} \label{sec:experiment}
The proposed approach was validated on a Clearpath Robotics Jackal and a Boston Dynamics SPOT quadruped, using MPC for the low-level controller, as it performed best in simulation. Since Spot can be approximated as a unicycle model with slower translational and rotational speed capabilities compared to the Jackal, the adaptation policy $\pi_\theta$ trained in simulation can be deployed on it. 

\begin{figure}[ht!]
     \vspace{-5pt}
    \centering
\includegraphics[width=.48\textwidth]{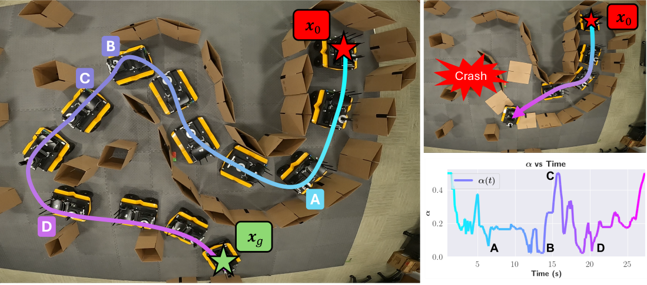}
       \vspace{-17pt}
    \caption{(Left) Jackal navigating through a cluttered environment with the full approach and (Top Right) crashing without CBF filter. (Bottom Right) shows $\alpha$ adaptation using our full approach.}
    \label{fig:jackal-experiments}
    \vspace{-15pt}
\end{figure}

Two test cases were setup for evaluation. In the first (Fig.~\ref{fig:jackal-experiments}), the Jackal navigates a snake-like path and forest environment toward a goal near the edge of the room. In the second (Fig.~\ref{fig:intro}), Spot negotiates an office environment where the goal is within a narrow corridor. Without the full approach, the low-level controller attempts to track $\bm{r}(\cdot)$ without considering safety, leading to a collision. Both figures show $\pi_\theta$ adapting $\alpha(t)$ as the vehicles navigate towards their final goals $\bm{x}_g$. 



\section{Conclusions and Future Work} \label{sec:conclusion} 
\vspace{-2pt}
In this work, we have presented a novel SAC-based adaptation scheme for the $\alpha$ parameter within the ZCBF safety constraint, enhancing the robustness of low-level control by enforcing safety while preventing deadlocks. Our approach has been exhaustively tested through simulation and experimental case studies. Additionally, it can be used with any low-level controller and system model with relative degree $1$ with respect to the ZCBF.

Future work will focus on detecting when the SAC policy encounters novel scenarios as the vehicle navigates, refining the policy in real-time through targeted simulations. We also aim to incorporate dynamic obstacles into our approach. 

\section{Acknowledgements}
This research is funded by the Commonwealth Cyber Initiative. The authors also thank Woosung Kim for assisting with experiments.
\newpage


\bibliographystyle{IEEEtran}
\bstctlcite{IEEEexample:BSTcontrol}
\bibliography{library}

\begin{thebibliography}{10}
\providecommand{\url}[1]{#1}
\csname url@samestyle\endcsname
\providecommand{\newblock}{\relax}
\providecommand{\bibinfo}[2]{#2}
\providecommand{\BIBentrySTDinterwordspacing}{\spaceskip=0pt\relax}
\providecommand{\BIBentryALTinterwordstretchfactor}{4}
\providecommand{\BIBentryALTinterwordspacing}{\spaceskip=\fontdimen2\font plus
\BIBentryALTinterwordstretchfactor\fontdimen3\font minus \fontdimen4\font\relax}
\providecommand{\BIBforeignlanguage}[2]{{%
\expandafter\ifx\csname l@#1\endcsname\relax
\typeout{** WARNING: IEEEtran.bst: No hyphenation pattern has been}%
\typeout{** loaded for the language `#1'. Using the pattern for}%
\typeout{** the default language instead.}%
\else
\language=\csname l@#1\endcsname
\fi
#2}}
\providecommand{\BIBdecl}{\relax}
\BIBdecl

\bibitem{nhtsa2024waymo}
\BIBentryALTinterwordspacing
{National Highway Traffic Safety Administration}, ``Waymo llc; recall 24e049; jaguar i-pace,'' National Highway Traffic Safety Administration, Tech. Rep. RCLRPT-24E049-1733, 2024. [Online]. Available: \url{https://static.nhtsa.gov/odi/rcl/2024/RCLRPT-24E049-1733.PDF}
\BIBentrySTDinterwordspacing

\bibitem{xiao2024barn}
\BIBentryALTinterwordspacing
X.~Xiao, Z.~Xu, A.~Datar, G.~Warnell, P.~Stone, J.~J. Damanik, J.~Jung, C.~A. Deresa, T.~D. Huy, C.~Jinyu, C.~Yichen, J.~A. Cahyono, J.~Wu, L.~Mo, M.~Lv, B.~Lan, Q.~Meng, W.~Tao, and L.~Cheng, ``Autonomous ground navigation in highly constrained spaces: Lessons learned from the 3rd barn challenge at icra 2024,'' 2024. [Online]. Available: \url{https://arxiv.org/abs/2407.01862}
\BIBentrySTDinterwordspacing

\bibitem{mohammad2024planner}
N.~Mohammad, J.~Higgins, and N.~Bezzo, ``A gp-based robust motion planning framework for agile autonomous robot navigation and recovery in unknown environments,'' in \emph{2024 IEEE International Conference on Robotics and Automation (ICRA)}, 2024, pp. 2418--2424.

\bibitem{ames2019cbf}
\BIBentryALTinterwordspacing
A.~D. Ames, S.~Coogan, M.~Egerstedt, G.~Notomista, K.~Sreenath, and P.~Tabuada, ``Control barrier functions: Theory and applications,'' 2019. [Online]. Available: \url{https://arxiv.org/abs/1903.11199}
\BIBentrySTDinterwordspacing

\bibitem{koushil2022polycbf}
A.~Thirugnanam, J.~Zeng, and K.~Sreenath, ``Safety-critical control and planning for obstacle avoidance between polytopes with control barrier functions,'' in \emph{2022 International Conference on Robotics and Automation (ICRA)}, 2022, pp. 286--292.

\bibitem{li2023cbf}
\BIBentryALTinterwordspacing
T.~Li and B.~Jayawardhana, ``Collision-free source seeking control methods for unicycle robots,'' 2023. [Online]. Available: \url{https://arxiv.org/abs/2212.07203}
\BIBentrySTDinterwordspacing

\bibitem{haarnoja2019sac}
\BIBentryALTinterwordspacing
T.~Haarnoja, A.~Zhou, K.~Hartikainen, G.~Tucker, S.~Ha, J.~Tan, V.~Kumar, H.~Zhu, A.~Gupta, P.~Abbeel, and S.~Levine, ``Soft actor-critic algorithms and applications,'' 2019. [Online]. Available: \url{https://arxiv.org/abs/1812.05905}
\BIBentrySTDinterwordspacing

\bibitem{brockman2016openaigym}
G.~Brockman, V.~Cheung, L.~Pettersson, J.~Schneider, J.~Schulman, J.~Tang, and W.~Zaremba, ``Openai gym,'' 2016.

\bibitem{ames2014firstcbf}
A.~D. Ames, J.~W. Grizzle, and P.~Tabuada, ``Control barrier function based quadratic programs with application to adaptive cruise control,'' in \emph{53rd IEEE Conference on Decision and Control}, 2014, pp. 6271--6278.

\bibitem{ames2017zcbf}
A.~D. Ames, X.~Xu, J.~W. Grizzle, and P.~Tabuada, ``Control barrier function based quadratic programs for safety critical systems,'' \emph{IEEE Transactions on Automatic Control}, vol.~62, no.~8, pp. 3861--3876, 2017.

\bibitem{nagumo1942invariance}
\BIBentryALTinterwordspacing
M.~Nagumo, ``{\"U}ber die lage der integralkurven gew{\"o}hnlicher differentialgleichungen,'' 1942. [Online]. Available: \url{https://api.semanticscholar.org/CorpusID:118866209}
\BIBentrySTDinterwordspacing

\bibitem{pen2023logitcbf}
C.~Peng, O.~Donca, G.~Castillo, and A.~Hereid, ``Safe bipedal path planning via control barrier functions for polynomial shape obstacles estimated using logistic regression,'' in \emph{2023 IEEE International Conference on Robotics and Automation (ICRA)}, 2023, pp. 3649--3655.

\bibitem{dai2022learncbf}
\BIBentryALTinterwordspacing
B.~Dai, P.~Krishnamurthy, and F.~Khorrami, ``Learning a better control barrier function,'' 2022. [Online]. Available: \url{https://arxiv.org/abs/2205.05429}
\BIBentrySTDinterwordspacing

\bibitem{ames2020adaptive}
A.~J. Taylor and A.~D. Ames, ``Adaptive safety with control barrier functions,'' in \emph{2020 American Control Conference (ACC)}, 2020, pp. 1399--1405.

\bibitem{lopez2020racbf}
B.~T. Lopez, J.-J.~E. Slotine, and J.~P. How, ``Robust adaptive control barrier functions: An adaptive and data-driven approach to safety,'' \emph{IEEE Control Systems Letters}, vol.~5, no.~3, pp. 1031--1036, 2021.

\bibitem{parwana2023rtcbf}
\BIBentryALTinterwordspacing
H.~Parwana and D.~Panagou, ``Rate-tunable control barrier functions: Methods and algorithms for online adaptation,'' 2023. [Online]. Available: \url{https://arxiv.org/abs/2303.12966}
\BIBentrySTDinterwordspacing

\bibitem{islam2015adaptive}
S.~Islam, M.~Faraz, R.~K. Ashour, J.~Dias, and L.~D. Seneviratne, ``Robust adaptive control of quadrotor unmanned aerial vehicle with uncertainty,'' in \emph{2015 IEEE International Conference on Robotics and Automation (ICRA)}, 2015, pp. 1704--1709.

\bibitem{fukuda2020cbftrajectory}
S.~Fukuda, Y.~Satoh, and O.~Sakata, ``Trajectory-tracking control considering obstacle avoidance by using control barrier function,'' in \emph{2020 International Automatic Control Conference (CACS)}, 2020, pp. 1--6.

\bibitem{chen2020fastrack}
M.~Chen, S.~L. Herbert, H.~Hu, Y.~Pu, J.~F. Fisac, S.~Bansal, S.~Han, and C.~J. Tomlin, ``Fastrack:a modular framework for real-time motion planning and guaranteed safe tracking,'' \emph{IEEE Transactions on Automatic Control}, vol.~66, no.~12, pp. 5861--5876, 2021.

\bibitem{perille2020barn}
D.~Perille, A.~Truong, X.~Xiao, and P.~Stone, ``Benchmarking metric ground navigation,'' in \emph{2020 IEEE International Symposium on Safety, Security, and Rescue Robotics (SSRR)}, 2020, pp. 116--121.

\bibitem{fossen2015mpc}
T.~I. Fossen, K.~Y. Pettersen, and R.~Galeazzi, ``Line-of-sight path following for dubins paths with adaptive sideslip compensation of drift forces,'' \emph{IEEE Transactions on Control Systems Technology}, vol.~23, no.~2, pp. 820--827, 2015.

\bibitem{gray2006geometry}
A.~Gray, S.~Salamon, and E.~Abbena, \emph{Modern Differential Geometry of Curves and Surfaces with Mathematica}.\hskip 1em plus 0.5em minus 0.4em\relax CRC Press, 2006, vol.~3, essay, n.d.

\bibitem{kraft1994slsqp}
D.~Kraft, ``Algorithm 733: {TOMP}--fortran modules for optimal control calculations,'' \emph{{ACM} Transactions on Mathematical Software}, vol.~20, pp. 262--281, 1994.

\bibitem{johnson2007nlopt}
S.~G. Johnson, ``The {NLopt} nonlinear-optimization package,'' \url{https://github.com/stevengj/nlopt}, 2007.

\bibitem{fox1997dwa}
D.~Fox, W.~Burgard, and S.~Thrun, ``The dynamic window approach to collision avoidance,'' \emph{IEEE Robotics \& Automation Magazine}, vol.~4, no.~1, pp. 23--33, 1997.

\bibitem{tordesillas2022faster}
J.~Tordesillas, B.~T. Lopez, M.~Everett, and J.~P. How, ``Faster: Fast and safe trajectory planner for navigation in unknown environments,'' \emph{IEEE Transactions on Robotics}, vol.~38, no.~2, pp. 922--938, 2022.

\end{thebibliography}

\end{document}